\documentclass[journal]{IEEEtran}

\ifCLASSINFOpdf
\else
   \usepackage[dvips]{graphicx}
\fi
\usepackage{url}

\usepackage{graphicx}

\usepackage{multirow}
\usepackage{adjustbox}
\usepackage{amsmath}
\usepackage{color}
\usepackage{amssymb}
\usepackage{hyperref}
\usepackage{booktabs}
\usepackage{makecell}
\usepackage{array}
\usepackage{caption}
\usepackage{subcaption}
\usepackage{tabularx}

\newcommand{\LRTwoFinalPhMAE}{30.19}
\newcommand{\LRTwoFinalPhACCt}{75.7}
\newcommand{\LRTwoFinalPB}{88.2}
\newcommand{\LRTwoFinalWordMAE}{39.9}
\newcommand{\LRTwoFinalWordACC}{84.5}
\newcommand{\LRTwoFinalWB}{83.9}
\newcommand{\LRThreeFinalPhMAE}{78.27}
\newcommand{\LRThreeFinalPhMAEt}{78.3}
\newcommand{\LRThreeFinalPhACC}{74.9}
\newcommand{\LRThreeFinalPB}{75.1}
\newcommand{\LRThreeFinalWordMAE}{91.8}
\newcommand{\LRThreeFinalWordACC}{86.4}
\newcommand{\LRThreeFinalWB}{64.8}
\newcommand{\BidirModelMAE}{30.68}

\newcommand{\OracleTwo}{28.47}
\newcommand{\OracleRecovered}{22}

\begin{document}

\title{Speaking-Clock Mamba: Phoneme-Conditioned State-Space Modeling for Visual Forced Alignment}

\author{Nan Lu, Qifan Liang

\thanks{Nan Lu and Qifan Liang are with the School of Computing, National University of Singapore, Singapore (e-mail: \mbox{nlu@u.nus.edu}; \mbox{liangqifan@u.nus.edu}). Nan Lu and Qifan Liang contributed equally to this work.}}

\markboth{Journal of \LaTeX\ Class Files, Vol. 14, No. 8, August 2015}
{Shell \MakeLowercase{\textit{et al.}}: Bare Demo of IEEEtran.cls for IEEE Journals}
\maketitle

\begin{abstract}
Visual forced alignment (VFA) locates phoneme and word boundaries in a silent talking-face video given its transcript. Existing methods mainly enforce transcript order at the decoding stage, leaving the encoder without an explicit mechanism to anchor its temporally evolving representation to transcript progress. As a result, phoneme-specific evidence is blurred and alignment drift accumulates over long sequences. In this paper, we propose Speaking-Clock Mamba, which makes transcript progress part of the temporal recurrence. A selective state-space (Mamba) backbone is inserted between the visual frontend and the Conformer cross-attention encoder, and a Phoneme-Conditioned Selective Scan (PCSS) conditions its input-dependent step size on a supervised, structurally monotone speaking clock that estimates progress through the transcript. Since causal and bidirectional variants of this backbone make complementary errors, a training-free selector picks the more confident alignment per utterance. Experiments on LRS2 and LRS3 show consistent gains over protocol-matched baselines, and the supervised clock provides an explicit estimate of alignment progress.
\end{abstract}


\IEEEpeerreviewmaketitle

\section{Introduction}

Visual forced alignment (VFA) estimates when each phoneme and word of a known transcript occurs in a silent talking-face video, without the waveform cues that acoustic forced alignment relies on. Its output feeds lip synchronization~\cite{syncnet,lip_sync}, audio-visual speech recognition~\cite{av_speech,autoavsr}, duration-controlled speech synthesis~\cite{tedtts}, and co-speech avatar animation~\cite{personagest}, where a transcript can be correct yet perceptually misaligned if its boundaries are systematically early, late, or unstable. Boundary precision is what a VFA system must deliver.

\begin{figure*}[t]
\centering
\includegraphics[width=0.9\textwidth]{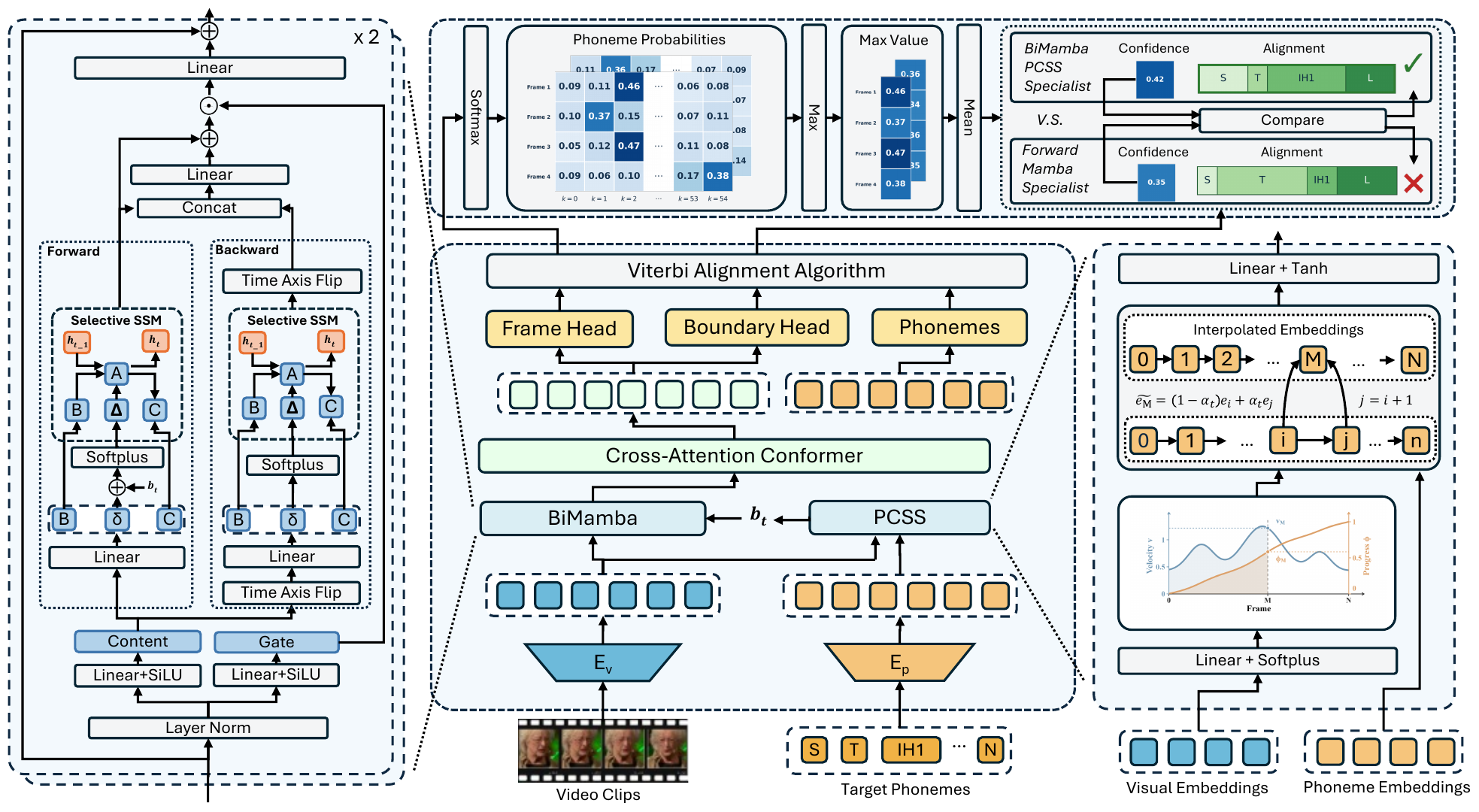}
\caption{Speaking-Clock Mamba. The clock reads a continuous target-phoneme location and biases the forward selective scan. A residual backward scan adds future context. The more confident of two decoded specialists is selected. Values in the top panel are illustrative.}
\label{fig:architecture}
\end{figure*}

Visible articulation makes this difficult. Acoustically distinct phonemes can share a mouth configuration~\cite{fisher}, coarticulation blurs boundaries in both directions~\cite{coarticulation}, and unlike keyword spotters~\cite{kwsnet,transpotter}, VFA must place every phoneme. Current systems pair a ResNet-based visual frontend~\cite{stafylakis} with an attention-based encoder~\cite{conformer} and impose order at the output, as in classical HMM forced alignment~\cite{rabiner}. DVFA retrieves transcript-relevant frames with multimodal attention~\cite{dvfa}, and CGL-Conformer strengthens local context and alignment supervision~\cite{cgl}. CTC~\cite{ctc}, soft-DTW~\cite{softdtw}, monotonic attention~\cite{monotonic}, position priors~\cite{ttsalign}, and Viterbi decoding~\cite{rabiner,cgl} all favor legal, ordered output paths. These designs share one limitation in that they enforce transcript order mainly at the decoding stage, leaving the encoder without an explicit mechanism to anchor its temporally evolving representation to transcript progress. Adjacent phonemes then compete for blurred evidence, alignment error accumulates over long sequences, and a decoder operating on the encoder's scores cannot undo either. Mechanisms that do track progress live elsewhere. For example, CIF accumulates evidence to emit tokens~\cite{cif}, monotonic attention constrains a decoder~\cite{monotonic}, monotonic alignment in TTS defines the path itself~\cite{ttsalign,glowtts}, and progress-gated conditioning in TTS acts at inference. None of them conditions the recurrence of the temporal model. Bidirectional and multi-directional selective scans exist for vision~\cite{vim,vmamba} and audio~\cite{audiomamba}, but their step size is driven by the input alone. This leaves the step size of a selective scan, the one quantity that sets how much of the past is kept, unconditioned on transcript progress.

This work studies selective state-space backbones for VFA under a controlled comparison and asks two questions under a fixed frontend and decoder. First, can a selective state-space (Mamba) backbone~\cite{mamba}, built on structured state spaces~\cite{s4}, inserted before the attention encoder, and equipped with an input-dependent step size that can express articulation pace, localize boundaries more precisely than attention alone? Second, does conditioning that step explicitly on transcript progress improve boundary precision, the accuracy of the progress estimate, or both? To this end, we propose Speaking-Clock Mamba, which inserts such a backbone before the encoder and conditions its step through a Phoneme-Conditioned Selective Scan (PCSS), in which a supervised, structurally monotone speaking clock accumulates framewise progress, reads the target-phoneme embedding at the clock position, and biases the step. A residual backward scan adds the anticipatory evidence that a causal scan cannot see. Because the causal and bidirectional models err on different utterances, we treat them as two specialists that are decoded separately, and one complete path is selected by posterior confidence, without a learned router or labels at inference.

On LRS2 and LRS3~\cite{lrs2,lrs3}, with Montreal Forced Aligner boundaries used only for supervision and scoring~\cite{mfa} and AV-HuBERT as a strong pretrained reference~\cite{avhubert}, the proposed system improves consistently over the Conformer-only pipeline, with gains that widen on long utterances, and reaches \LRTwoFinalPhMAE\,ms phoneme MAE on LRS2 and \LRThreeFinalPhMAE\,ms on a protocol-matched LRS3 subset, against 92.61\,ms for a reproduced CGL-Conformer. The supervised clock additionally provides an inspectable estimate of transcript progress.

\section{Proposed Method}

\subsection{Problem Formulation}
\label{sec:backbone}

Let $\mathbf V=[\mathbf v_1,\ldots,\mathbf v_T]$ be a silent visual sequence of $T$ frames and $\mathbf p=(p_1,\ldots,p_N)$ its transcript of $N$ phonemes. VFA seeks a monotone path $\mathbf a=(a_1,\ldots,a_T)$ that assigns every frame a phoneme index $a_t$, starts at the first phoneme, ends at the last, and advances by at most one phoneme per frame.

Figure~\ref{fig:architecture} shows the pipeline in its center panel, with PCSS on the right, BiMamba on the left, and the specialist selection at the top. A 3D-convolution and ResNet-18 frontend yields $\mathbf V$, and the proposed backbone, inserted here, transforms it. A four-layer Conformer cross-attention encoder fuses the result with the transcript embeddings and feeds three heads, which output the frame posteriors $\boldsymbol\pi_t$ over the phoneme classes, boundary scores, and an autoregressive phoneme sequence. A Viterbi decoder recovers $\mathbf a$ from $\boldsymbol\pi_t$, with a boundary score above a threshold forcing an advance.

\subsection{PCSS: Phoneme-Conditioned Selective Scan}
\label{sec:pcss}

PCSS turns an estimate of transcript progress into a per-frame signal that Sec.~\ref{sec:bidir} injects into the step size of the selective scan.

\noindent\textbf{Speaking clock.}
A linear layer with a softplus, plus a small constant, predicts a positive rate $r_t$ for every frame from $\mathbf v_t$. The clock is the normalized cumulative rate
\begin{equation}
\phi_t=\frac{\sum_{j\le t}r_j}{\sum_{j\le T}r_j}.
\label{eq:clock}
\end{equation}
Thus $\phi_t$ is monotone by construction rather than by a penalty, identical at training and inference, and equal to $t/T$ at initialization. Because the normalization spans the utterance, the forward scan is offline.
\begin{table*}[!t]
\centering
\caption{Results on LRS2 and LRS3. Results for KWS-Net–DVFA are quoted from~\cite{dvfa}, and CGL-Conformer is reproduced from the released code using our data and labels. On LRS3, Ours denotes the BiMamba PCSS specialist. Bold and underlined values are the best and second-best results, and a dash marks results that are not reported or not applicable.}
\label{tab:comparison}
\small
\setlength{\tabcolsep}{1mm}
\begin{tabular*}{\textwidth}{
@{\extracolsep{\fill}}l ccc ccc ccc ccc@{}
}
\toprule
& \multicolumn{6}{c}{\textbf{LRS2}}
& \multicolumn{6}{c}{\textbf{LRS3}} \\
\cmidrule(lr){2-7}
\cmidrule(lr){8-13}
& \multicolumn{3}{c}{\textbf{Phoneme}}
& \multicolumn{3}{c}{\textbf{Word}}
& \multicolumn{3}{c}{\textbf{Phoneme}}
& \multicolumn{3}{c}{\textbf{Word}} \\
\cmidrule(lr){2-4}
\cmidrule(lr){5-7}
\cmidrule(lr){8-10}
\cmidrule(lr){11-13}
\textbf{Method}
& MAE$\downarrow$ & ACC$\uparrow$ & PB@50$\uparrow$
& MAE$\downarrow$ & ACC$\uparrow$ & WB@50$\uparrow$
& MAE$\downarrow$ & ACC$\uparrow$ & PB@50$\uparrow$
& MAE$\downarrow$ & ACC$\uparrow$ & WB@50$\uparrow$ \\
\midrule
KWS-Net~\cite{kwsnet}
& -- & -- & --
& 171.5 & 53.0 & --
& -- & -- & --
& 262.9 & 42.6 & -- \\
CTC-based~\cite{dvfa}
& -- & -- & --
& 80.4 & 71.7 & --
& -- & -- & --
& 124.5 & 60.1 & -- \\
Transpotter~\cite{transpotter}
& -- & -- & --
& 71.7 & 75.0 & --
& -- & -- & --
& 167.3 & 61.8 & -- \\
DVFA~\cite{dvfa}
& 176.5 & 61.3 & --
& 67.7 & \underline{84.2} & --
& 249.1 & 57.1 & --
& \underline{97.7} & \underline{80.2} & -- \\
CGL-Conformer~\cite{cgl}
& 41.2 & \underline{71.4} & 84.1
& 50.2 & 83.2 & 80.2
& 92.61 & \underline{73.0} & 72.9
& 131.9 & 65.8 & 63.8 \\
AV-HuBERT~\cite{avhubert}
& \underline{30.78} & 62.4 & \underline{86.2}
& \underline{44.2} & 82.8 & \underline{81.7}
& \underline{79.4} & 64.9 & \textbf{79.1}
& 130.1 & 76.4 & \underline{64.1} \\
\textbf{Ours}
& \textbf{\LRTwoFinalPhMAE} & \textbf{\LRTwoFinalPhACCt} & \textbf{\LRTwoFinalPB}
& \textbf{\LRTwoFinalWordMAE} & \textbf{\LRTwoFinalWordACC} & \textbf{\LRTwoFinalWB}
& \textbf{\LRThreeFinalPhMAEt} & \textbf{\LRThreeFinalPhACC} & \underline{\LRThreeFinalPB}
& \textbf{\LRThreeFinalWordMAE} & \textbf{\LRThreeFinalWordACC} & \textbf{\LRThreeFinalWB} \\
\bottomrule
\end{tabular*}
\end{table*}

\noindent\textbf{Continuous phoneme readout.}
The transcript embeddings, with BOS and EOS tokens added, are placed on a uniform grid over the unit interval. The readout $\widetilde{\mathbf e}_t$ is the linear interpolation of the two grid embeddings that bracket the clock position $\phi_t$, so frames near a boundary read a blend of the adjacent phonemes. One clock is shared by all layers.

\noindent\textbf{Step bias.}
The readout is projected and squashed into a bounded bias
\begin{equation}
\mathbf b_t=\tau\tanh(\mathbf W_{\rm ph}\widetilde{\mathbf e}_t+\mathbf c_{\rm ph}),
\label{eq:bias}
\end{equation}
with $\tau=0.5$. $\mathbf W_{\rm ph}$ is initialized from a zero-mean Gaussian and $\mathbf c_{\rm ph}$ at zero, and the $\tanh$ bounds each entry of $\mathbf b_t$ by $\tau$.

\noindent\textbf{Clock supervision.}
The clock is supervised towards a coarse monotone progress target, defined as the number of ground-truth label changes up to frame $t$, including those into and out of silence, divided by $N+2$ and capped at one, where $N+2$ is the transcript length with BOS and EOS. As this target is only a surrogate, the clock loss of Sec.~\ref{sec:objective} is not expected to reach zero. The clock therefore has a direct interpretation as normalized transcript progress rather than an unconstrained latent gate. Its supervision encourages the recurrence to associate visual time with a monotone location in the known transcript.

\subsection{BiMamba: Residual Bidirectional Scan}
\label{sec:bidir}

\noindent\textbf{Selective scan.}
For a layer input $\mathbf x_t$, a selective scan updates a hidden state $\mathbf h_t$ and reads it out as
\begin{equation}
\mathbf h_t=\overline{\mathbf A}(\Delta_t)\mathbf h_{t-1}
+\overline{\mathbf B}(\Delta_t)\mathbf x_t,
\qquad \mathbf y_t=\mathbf C_t\mathbf h_t+\mathbf D\mathbf x_t,
\label{eq:ssm}
\end{equation}
with discretized matrices that depend on a positive step $\Delta_t$. Mamba derives the step from the input alone as $\Delta_t=\operatorname{softplus}(\delta_t)$~\cite{mamba}. A large step drives $\overline{\mathbf A}$ towards zero and resets the state, as at a transition, while a small step preserves it, as within a sustained articulation. The step thus expresses articulation pace, and one causal layer of this scan is the Forward Mamba backbone.

\noindent\textbf{Clock-conditioned step.}
PCSS adds the bias of Eq.~\ref{eq:bias} to the step before the softplus,
\begin{equation}
\Delta_t=\operatorname{softplus}(\delta_t+\mathbf b_t).
\label{eq:pcss}
\end{equation}
Placing the signal on the step modulates the dynamics of the recurrence rather than the content of the frame. Both $\delta_t$ and $\mathbf b_t$ have one entry per channel, and the softplus acts elementwise. With this step, Forward Mamba becomes Forward Mamba PCSS.

\noindent\textbf{Residual backward scan.}
A causal scan cannot use future visual evidence, which is informative near ambiguous phoneme boundaries. We therefore add a backward scan. Each BiMamba layer (Fig.~\ref{fig:architecture}, left) splits its normalized input into a content branch $\mathbf U$ and a gate branch $\mathbf Z$. Running the scan of Eq.~\ref{eq:ssm} forward over $\mathbf U$ with the step of Eq.~\ref{eq:pcss} gives $\mathbf Y_{\rm f}$. A backward scan with its own state-space parameters, run on the time-flipped sequence and flipped back as in bidirectional visual SSMs~\cite{vim}, gives $\mathbf Y_{\rm b}$. They are fused by a learned residual,
\begin{equation}
\mathbf R=\mathbf Y_{\rm f}+\mathbf W_{f}[\mathbf Y_{\rm f};\mathbf Y_{\rm b}],
\label{eq:bidir}
\end{equation}
and $\mathbf R$, gated by $\operatorname{SiLU}(\mathbf Z)$ and projected, is added to the layer input as in a standard Mamba block. The residual keeps an unmodified forward path while $\mathbf W_f$ injects backward context channel by channel, and $\mathbf W_f$ is initialized with small values so that BiMamba starts close to the forward path. Only the forward scan is clock-conditioned, so that BiMamba PCSS shares the conditioned path of Forward Mamba PCSS and the backward scan contributes unconditioned future context. We call the resulting two-layer bidirectional backbone BiMamba and, with the clock-conditioned forward scan, BiMamba PCSS.

\subsection{Confidence-Based Specialist Selection}
\label{sec:inference}

Inspired by~\cite{stanet}, the full system combines two separately trained specialists at inference, because causal and bidirectional models have complementary inductive biases (causal precision versus anticipatory context) and, empirically, complementary utterance-level errors (Sec.~\ref{sec:selector_analysis}). We therefore pair Forward Mamba with BiMamba PCSS, decode each on its own, and select one complete alignment per utterance. The confidence of a specialist is the mean over frames of its maximum frame posterior,
\begin{equation}
c=\frac{1}{T}\sum_{t=1}^{T}\max_k\pi_{t,k},
\label{eq:selector}
\end{equation}
with $\boldsymbol\pi_t$ the posteriors of Sec.~\ref{sec:backbone} produced by that specialist, which measures how far it has committed to a segmentation. The specialist with the larger confidence supplies the complete path (Fig.~\ref{fig:architecture}, top). We select rather than average, because averaging two sharp but offset boundary distributions can blur them and move the Viterbi path between the two, whereas selection keeps one self-consistent path (compared in Sec.~\ref{sec:selector_analysis}). The selector needs neither labels nor a learned router.

\subsection{Training Objective}
\label{sec:objective}

Training minimizes
\begin{equation}
\mathcal L=\mathcal L_{\rm frame}+\mathcal L_{\rm phon}+\mathcal L_{\rm bound}
+\lambda_{\rm SV}\mathcal L_{\rm SV}+\lambda_{\rm clk}\mathcal L_{\rm clk}.
\label{eq:objective}
\end{equation}
The first three terms supervise the heads of Sec.~\ref{sec:backbone}, where
$\mathcal L_{\rm frame}$ and $\mathcal L_{\rm phon}$ are the cross-entropies of the frame and sequence heads, and $\mathcal L_{\rm bound}$ is the binary cross-entropy of the boundary head. The alignment loss $\mathcal L_{\rm SV}$ is the Viterbi cost of the transcript with BOS and EOS, with the minimum over staying and advancing replaced by the soft-min of soft-DTW~\cite{softdtw,diffdp}. It is enabled after five warm-up epochs so that the heads stabilize first. The clock loss $\mathcal L_{\rm clk}$ is the mean squared error between the clock of Eq.~\ref{eq:clock} and its target of Sec.~\ref{sec:pcss} over valid frames.

\section{Experiments}
\label{sec:experiments}

\subsection{Protocol}

\noindent\textbf{Datasets and metrics.}
We use LRS2, BBC speech with 47,198 training and 963 aligned test utterances, as the primary corpus, and LRS3, TED speech with about 40\,k training and 5\,k test utterances, on which models are trained from an LRS2 warm start. Montreal Forced Aligner~\cite{mfa} derives phoneme and word targets from the audio track. They are used only for supervision and scoring, and at inference the model receives only the silent video and its transcript. The frame head predicts over a 55-token vocabulary. Following prior VFA work~\cite{dvfa,cgl}, we report utterance-averaged boundary MAE in ms and frame ACC. In addition, PB@50 and WB@50 give the percentages of phoneme and word boundaries within 50\,ms of the reference.

\noindent\textbf{Implementation details.}
A 3D-convolution and ResNet-18 frontend maps grayscale 25\,fps video to 512-dimensional per-frame features and is fine-tuned end-to-end from pretrained weights.
Models use Adam with $\beta_1=0.9$ and $\beta_2=0.997$, batch size 8, weight decay $10^{-4}$, and a Noam schedule with 8 warmup epochs followed by $\mathrm{step}^{-1/2}$ decay. Mamba uses a state dimension of 16 and S4D-real initialization~\cite{s4d}. The Forward models use one Mamba layer and BiMamba two. Models are warm-started from the corresponding converged backbone checkpoints and trained for 30–60 epochs, depending on the configuration. Checkpoints are saved periodically during training. For evaluation, we apply stochastic weight averaging (SWA)~\cite{swa} over multiple saved checkpoints and report the performance of the resulting averaged model, rather than selecting the best individual checkpoint. All experiments are run on four NVIDIA RTX A5000 GPUs using PyTorch 2.0.1.

\subsection{Comparison with Prior Work}

Table~\ref{tab:comparison} compares with prior VFA systems that align the transcript to the silent video directly. Pipelines that synthesize speech and align it with an audio aligner, such as AlignDiT~\cite{aligndit}, lie outside our controlled comparison. Rows KWS-Net to DVFA are quoted as published. CGL-Conformer is reproduced on both corpora from the authors' released code with our data and labels, and AV-HuBERT~\cite{avhubert} feeds frozen features to a VFA head trained on the target corpus, its LRS3 pretraining making those results in-domain. On LRS2, our system leads all six columns. It is 0.6 and 4.3\,ms below AV-HuBERT in phoneme and word MAE, about two points above it in hit rate, and highest in frame accuracy at both levels, 4.3 points above CGL-Conformer for phonemes and 0.3 above DVFA for words. AV-HuBERT's frozen features give competitive timing but frame labels 13 points less accurate than ours. On LRS3, published results come from different pipelines and data splits. Compared with CGL-Conformer reproduced using the same data and labels, the BiMamba PCSS specialist reduces phoneme MAE by 14.3\,ms and improves all three word-level metrics.

\subsection{Ablation Studies}
\label{sec:ablation}

\begin{table}[!htb]
\centering
\caption{Ablation on LRS2 at the phoneme level. Indented rows are variants of Forward Mamba. Bold and underlined values are the best and second-best results.}
\label{tab:ablation}
\small
\begin{tabular*}{\columnwidth}{@{\extracolsep{\fill}}lcc@{}}
\toprule
Configuration & MAE$\downarrow$ & ACC$\uparrow$ \\
\midrule
Baseline & 32.75 & 74.96 \\
+ Forward Mamba & 30.86 & 75.41 \\
\quad + second causal layer & 31.19 & \underline{75.56} \\
\quad + PCSS, clock fixed to $t/T$ & 30.94 & 75.50 \\
\quad + PCSS (Forward Mamba PCSS) & 30.48 & 75.42 \\
+ selection with Forward Mamba PCSS & \underline{30.36} & 75.55 \\
+ selection with BiMamba PCSS (ours) & \textbf{30.19} & \textbf{75.65} \\
\bottomrule
\end{tabular*}
\end{table}

\begin{table}[!htb]
\centering
\caption{Training objective of Forward Mamba PCSS on LRS2.}
\label{tab:objective}
\small
\begin{tabular*}{\columnwidth}{@{\extracolsep{\fill}}lccc@{}}
\toprule
Objective & MAE$\downarrow$ & ACC$\uparrow$ & PB@50$\uparrow$ \\
\midrule
$\mathcal L_{\rm frame}+\mathcal L_{\rm phon}+\mathcal L_{\rm bound}$ & 31.19 & 73.90 & 88.3 \\
+ $\mathcal L_{\rm SV}$ & 30.57 & 74.55 & 88.0 \\
+ $\mathcal L_{\rm SV}$ + $\mathcal L_{\rm clk}$ (ours) & \textbf{30.48} & \textbf{75.42} & \textbf{88.9} \\
\bottomrule
\end{tabular*}
\end{table}

Table~\ref{tab:ablation} builds the system up from the Conformer-only baseline. Forward Mamba gives the largest gain, 1.9\,ms. A second causal layer or a clock fixed to $t/T$ does not improve on it, whereas the learned, supervised clock makes Forward Mamba PCSS the best single model, 0.4\,ms below Forward Mamba. In Table~\ref{tab:objective}, $\mathcal L_{\rm SV}$ accounts for 0.6 of the 0.7\,ms gap to the full objective and $\mathcal L_{\rm clk}$ for the rest. Clock supervision lowers MAE only marginally, from 30.57 to 30.48\,ms, while reducing the clock's mean deviation from the target from 6.7 to 5.9 percentage points of progress. BiMamba PCSS is the better selection partner, 0.2\,ms below the Forward Mamba PCSS pairing. Over the chain, MAE falls by 2.6\,ms while ACC moves by less than one point, indicating that the components sharpen boundaries rather than frame labels. Forward Mamba, Forward Mamba PCSS, and BiMamba PCSS add 2.7, 6.4, and 12.3\,M parameters to the 62.5\,M baseline, and selection costs one extra forward pass.

\subsection{Effect of Specialist Selection}
\label{sec:selector_analysis}

Table~\ref{tab:selector_ablation} compares selection strategies over the same per-utterance predictions of the two specialists. Boundary error is heavy-tailed, the worst 10\% of utterances carrying 40.3\% of the total, and the oracle shows 2.2\,ms of headroom below the stronger specialist. Max-posterior routing gains 0.5\,ms over the stronger specialist, 0.6\,ms over random and 0.2\,ms over entropy routing, closes \OracleRecovered\% of the oracle gap, and matches the stronger specialist on the hardest decile. Averaging the two specialists' logits or posteriors before decoding is 0.7 and 0.3\,ms worse than selection.

\begin{table}[!htb]
\centering
\caption{Specialist selection on LRS2. Hard is the MAE on the 10\% of utterances with the largest Forward Mamba error. Oracle uses ground-truth error.}
\label{tab:selector_ablation}
\small
\setlength{\tabcolsep}{2pt}
\begin{tabular*}{\columnwidth}{@{\extracolsep{\fill}}lccc@{}}
\toprule
\textbf{Strategy} & \textbf{MAE}$\downarrow$ & \textbf{PB@50}$\uparrow$ & \textbf{Hard}$\downarrow$ \\
\midrule
Forward Mamba specialist & 30.86 & 87.9 & 124.76 \\
BiMamba PCSS specialist & \BidirModelMAE & 87.9 & \textbf{117.41} \\
\midrule
Random routing & 30.77 & \underline{88.1} & 121.07 \\
Entropy routing & \underline{30.40} & 87.1 & 118.14 \\
\textbf{Max-posterior routing (ours)} & \textbf{\LRTwoFinalPhMAE} & \textbf{\LRTwoFinalPB} & \underline{117.51} \\
\midrule
Oracle upper bound & \OracleTwo & -- & -- \\
\bottomrule
\end{tabular*}
\end{table}

\subsection{Effect of Phoneme Class and Utterance Length}
\label{sec:length}

\begin{table}[!htb]
\centering
\caption{LRS2 phoneme boundary MAE in ms by class of the reference phoneme, for the Conformer-only baseline and the BiMamba PCSS specialist.}
\label{tab:phoneme_class}
\small
\setlength{\tabcolsep}{2pt}
\begin{tabular*}{\columnwidth}{@{\extracolsep{\fill}}lcccccc@{}}
\toprule
Class & Vowel & Plosive & Fricative & Affricate & Nasal & Approx. \\
\midrule
Baseline & 34.0 & 33.3 & 35.0 & 37.7 & 31.9 & 34.1 \\
Ours & \textbf{31.7} & \textbf{30.5} & \textbf{34.1} & \textbf{33.1} & \textbf{29.8} & \textbf{30.8} \\
\bottomrule
\end{tabular*}
\end{table}

\begin{table}[!htb]
\centering
\caption{LRS2 phoneme MAE in ms by utterance length in phonemes, for the Conformer-only baseline and our full system. The bins hold 208, 270, 269, and 216 utterances.}
\label{tab:length}
\small
\setlength{\tabcolsep}{5pt}
\begin{tabular}{@{}lcccc@{}}
\toprule
Length & 8--13 & 14--19 & 20--29 & $\geq$30 \\
\midrule
Baseline & 26.6 & 34.0 & 31.7 & 38.3 \\
Ours & \textbf{26.1} & \textbf{31.0} & \textbf{29.2} & \textbf{34.3} \\
\bottomrule
\end{tabular}
\end{table}

Table~\ref{tab:phoneme_class} breaks LRS2 phoneme MAE down by class of the reference phoneme. The BiMamba PCSS specialist lowers MAE in every class, by 7.7\% on average. The largest reduction, 12.2\%, falls on affricates, the class the baseline finds hardest, followed by approximants and plosives, the classes whose visual evidence is weakest and depends most on context. The smallest, 2.6\%, falls on fricatives, whose sustained constriction the baseline already localizes well. The spread across classes narrows by a quarter, and fricatives replace affricates as the hardest class. Table~\ref{tab:length} breaks it down by utterance length. The relative gain of the full system over the baseline rises from 1.9\% on the shortest bin to 10.4\% on the longest, and the rise in MAE from the shortest to the longest bin is 30\% smaller for our system than for the baseline, so the benefit grows with utterance length. The gain becomes larger on longer utterances, consistent with transcript-progress conditioning helping reduce alignment drift over long sequences.

\section{Conclusion}

In this paper, we propose Speaking-Clock Mamba, a selective state-space framework for visual forced alignment that makes transcript progress part of the temporal recurrence through a phoneme-conditioned selective scan driven by a supervised monotone speaking clock, complemented by a residual bidirectional scan and confidence-based specialist selection. Extensive experiments demonstrate improved phoneme-level boundary accuracy on LRS2 and LRS3, increasing gains with utterance length, and an explicit estimate of alignment progress.

\end{document}